\documentclass[11pt, a4paper, logo, copyright, nonumbering]{kwaipilot}
\usepackage{dblfloatfix}
\usepackage{ulem}
\usepackage{caption}
\usepackage{dramatist}
\usepackage{xspace}
\usepackage{pifont} %
\usepackage{multirow}
\usepackage{tcolorbox}
\usepackage{xltabular}
\usepackage{longtable}
\usepackage{hyperref}
\usepackage{graphicx}
\usepackage{subcaption}
\usepackage{float}
\usepackage{amsfonts}
\usepackage{amsmath}
\usepackage{amssymb}
\usepackage{lineno}
\usepackage{multirow}
\usepackage{adjustbox}

\usepackage[bottom]{footmisc}

\usepackage{CJKutf8}
\usepackage{setspace}

\usepackage{dsfont}
\usepackage{array} %
\usepackage{tabularx} %
\usepackage{xcolor} %
\usepackage{tabularx}
\usepackage{booktabs}

\usepackage{lipsum}  %
\usepackage{multicol} %

\usepackage{indentfirst} 
\usepackage[sort&compress,numbers]{natbib}

\makeatletter
\def\@BTrule[#1]{%
  \ifx\longtable\undefined
    \let\@BTswitch\@BTnormal
  \else\ifx\hline\LT@hline
    \nobreak
    \let\@BTswitch\@BLTrule
  \else
     \let\@BTswitch\@BTnormal
  \fi\fi
  \global\@thisrulewidth=#1\relax
  \ifnum\@thisruleclass=\tw@\vskip\@aboverulesep\else
  \ifnum\@lastruleclass=\z@\vskip\@aboverulesep\else
  \ifnum\@lastruleclass=\@ne\vskip\doublerulesep\fi\fi\fi
  \@BTswitch}
\makeatother

\addto\extrasenglish{
}

 {\begin{list}{}%
         {\setlength{\leftmargin}{#1}}%
         \item[]%
 }
 {\end{list}}
 
\reportnumber{001} %

\renewcommand{\today}{}

\title{\centering SRPO: A Cross-Domain Implementation of Large-Scale Reinforcement Learning on LLM}

\author[*]{
Kuaishou Kwaipilot Team\\
\small \normalfont HuggingFace: \url{https://huggingface.co/Kwaipilot/SRPO-Qwen-32B/}
}

\renewcommand{\phi}{\varphi}

\renewcommand{\epsilon}{\varepsilon}
\renewcommand{\imath}{\mathrm{i}}

\newlength{\restsubwidth}
\newlength{\restsubheight}
\newlength{\restsubmoreheight}
\newcommand{\rest}[2]{%
        \settowidth{\restsubwidth}{\ensuremath{#2}}
        \settoheight{\restsubheight}{\ensuremath{{}_{#2}}}
        \ensuremath{{#1\hskip 0.5pt}_{\vrule\kern2pt\parbox[b][%
        4pt][b]{\the\restsubwidth}{%
                        \ensuremath{{}_{#2}}}}}
        }

\begin{abstract}
\footnotetext{Model available at: \url{https://huggingface.co/Kwaipilot/SRPO-Qwen-32B/}}

Recent advances of reasoning models, exemplified by OpenAI's o1 and DeepSeek's R1, highlight the significant potential of Reinforcement Learning (RL) to enhance the reasoning capabilities of Large Language Models (LLMs). However, replicating these advancements across diverse domains remains challenging due to limited methodological transparency. In this work, we present two-Staged history-Resampling Policy Optimization (SRPO), which surpasses the performance of DeepSeek-R1-Zero-32B on the AIME24 and LiveCodeBench benchmarks. SRPO achieves this using the same base model as DeepSeek (i.e. Qwen2.5-32B), using only about 1/10 of the training steps required by DeepSeek-R1-Zero-32B, demonstrating superior efficiency. Building upon Group Relative Policy Optimization (GRPO), we introduce two key methodological innovations: (1) a two-stage cross-domain training paradigm designed to balance the development of mathematical reasoning and coding proficiency, and (2) History Resampling (HR), a technique to address ineffective samples. Our comprehensive experiments validate the effectiveness of our approach, offering valuable insights into scaling LLM reasoning capabilities across diverse tasks. 

\end{abstract}

\begin{document}

\maketitle

\begin{figure}[h]
\centering
\begin{minipage}{0.46\textwidth}
    \centering
    \includegraphics[width=\textwidth]{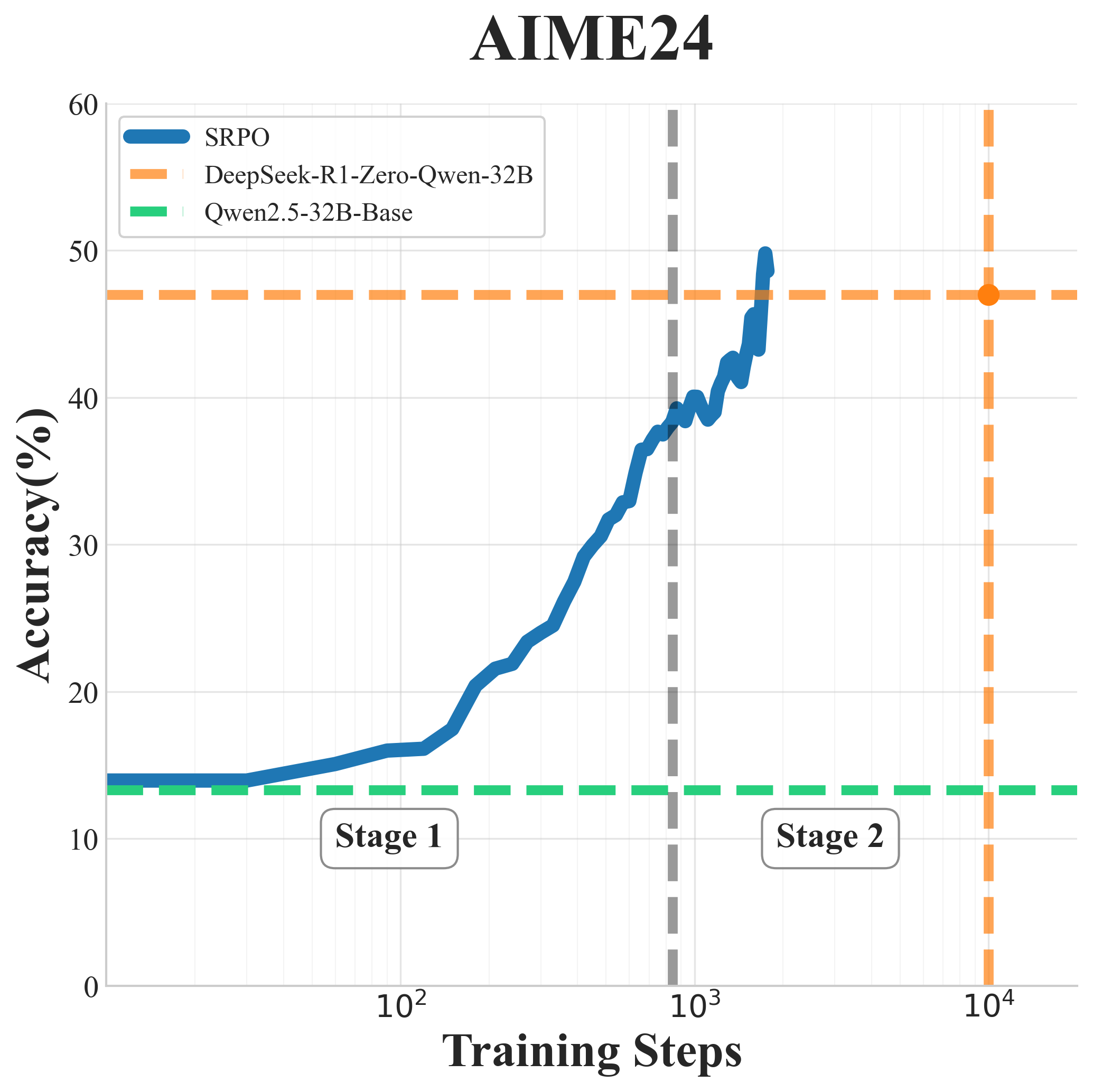}
\end{minipage}%
\hfill
\begin{minipage}{0.46\textwidth}
    \centering
    \includegraphics[width=\textwidth]{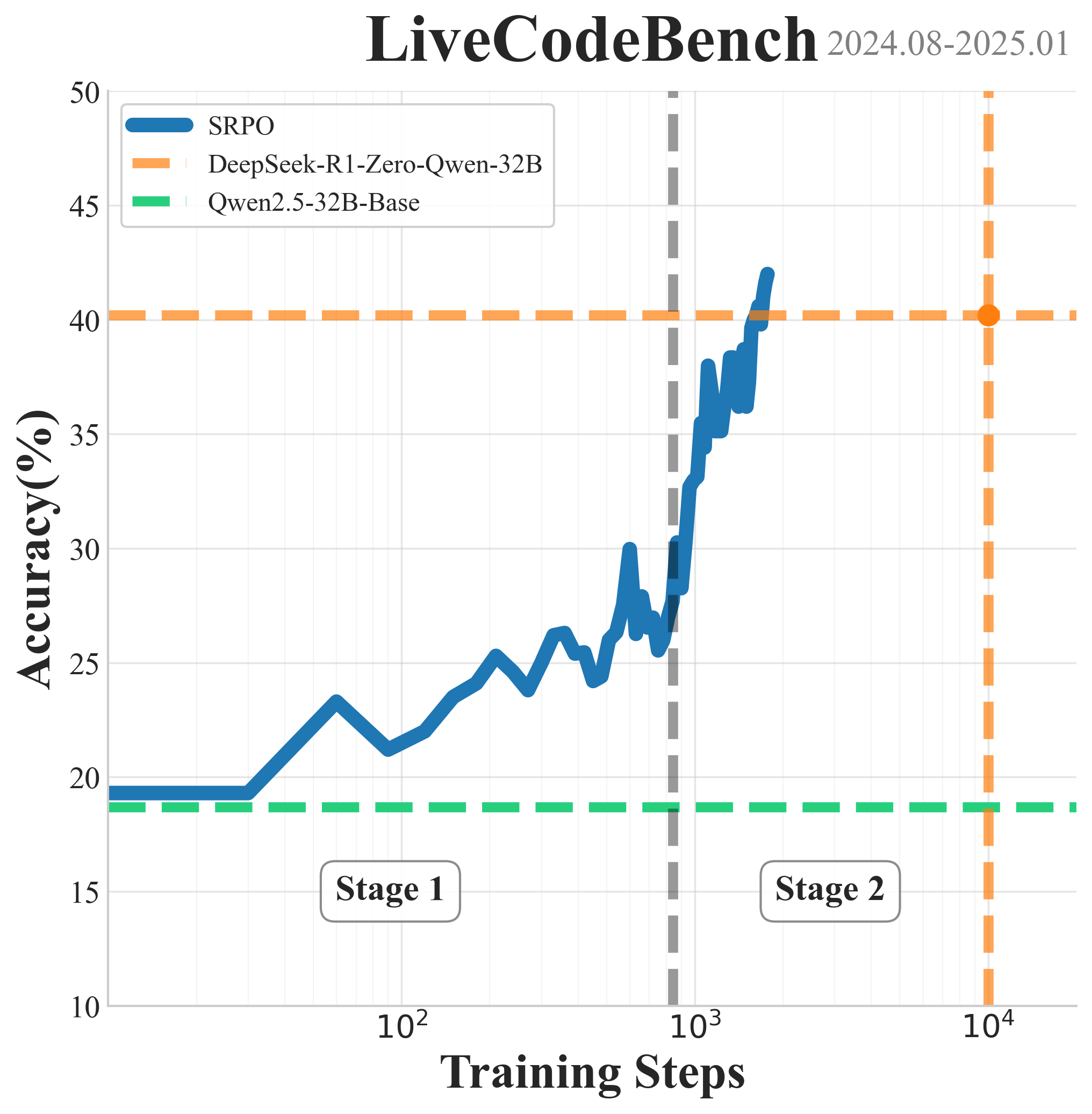}
\end{minipage}
\caption{\raggedright SRPO achieves superior results with only 10\% of DeepSeek's training steps. The values shown are pass@1 scores, averaged over 32 samples per question.}
\label{fig:benchmark_performance}
\end{figure}

\newpage

\section{Introduction}

The success of OpenAI O1-series \cite{openai2024reason, openai2024o1} and DeepSeek-R1 \cite{deepseek2025r1} demonstrates that large-scale Reinforcement Learning has become a highly effective approach to elicit complex reasoning behaviors and enhance the capabilities of Large Language Models (LLMs) \cite{openai2023gpt4,anthropic2024claude, deepseek2024v3}. However, the core training methodologies of these leading reasoning models remain largely undisclosed in their technical reports \cite{openai2024o1,deepseek2025r1,xai2024grok,deepmind2024gemini,qwen2024qwq,du2025kimi}. Recently, Open-Reasoner-Zero \cite{chen2025empirical} and DAPO \cite{yu2025dapo} provided detailed reproduction reports, but their work exclusively addresses mathematical reasoning, leaving the broader challenge of cross-domain generalization insufficiently explored. Furthermore, commonly observed GRPO \cite{shao2024deepseekmath} training issues, such as performance plateaus, sample inefficiency, and inadequate development of specialized reasoning skills when training in mixed-domain datasets, further complicate the effective scaling of RL methods.

In this work, we propose two-\textbf{S}taged history-\textbf{R}esampling \textbf{P}olicy \textbf{O}ptimization (\textbf{SRPO}), a novel RL framework designed to systematically address the above training challenges from multiple perspectives, achieving superior results across both mathematical and coding domains.

Specifically, we implement a two-stage training paradigm to develop reasoning and domain-specific skills in LLMs. We find that enhancing long Chain-of-Thought (CoT) reasoning ability early on is essential for generalizing across tasks. In the first stage, we train primarily on mathematical data to foster reflective thinking and step-by-step problem-solving. In the second stage, we integrate coding data, building on the foundation of reasoning skills developed in the first stage. This paradigm ensures steady improvement in both reasoning and coding abilities.

To address the zero-advantage phenomenon in GRPO, which hampers gradient updates and reduces sample efficiency, we introduce history resampling. By filtering out consistently correct answers, we ensure meaningful gradients, improve sample efficiency, and accelerate convergence.

Furthermore, we detail our data curation pipeline, which includes data cleaning and difficulty-level categorization. The final training dataset exposes the model to challenging, high-quality problems designed to foster sophisticated reasoning abilities.

Combining the aforementioned techniques, SRPO achieves 50.0 pass@1 on AIME24 and 41.6 pass@1 on LiveCodeBench \cite{jain2024livecodebench}, surpassing the performance of DeepSeek-R1-Zero-Qwen-32B \cite{deepseek2025r1} (47.0 points on AIME \& 40.2 on LiveCodeBench), training for over 1K steps --- just about 1/10 of the R1-Zero training steps. In the latter part of the article, we provide a detailed analysis of the reasoning patterns in large-scale RL training on math and coding tasks.

In summary, our contributions are threefold:
\begin{itemize}
    \item We propose SRPO, the first algorithm fully reproducing DeepSeek-R1's performance on both math and code tasks, with detailed technical analysis.
    \item We introduce the two-stage training paradigm, history resampling, and a dedicated math \& code RL data curation pipeline.
    \item We provide an in-depth analysis of the challenges and thinking behavior in large-scale RL training for math and code, offering empirical insights for future research.
\end{itemize}

\newpage

\section{Preliminary}
\subsection{GRPO}
The core idea of GRPO \cite{shao2024deepseekmath} is to estimate the baseline through a relative reward within a group of rollouts. This approach reduces the computational cost of the critic model and improves the training stability. Specifically, for each question q, the model generates a group of responses $o_1, o_2,... , o_G$ and calculates the corresponding rewards $r_1, r_2, ... , r_G$. $A_i$ is the advantage obtained by normalizing the rewards within each group.
\begin{equation}
\begin{split}
    \mathcal{J}_{GRPO}(\theta) &= \mathbb{E}{[q \sim P(Q), \{o_i\}_{i=1}^G \sim \pi_{\theta_{old}}(O|q)]}  \\
    & \frac{1}{G}\sum_{i=1}^G \left( \min \left( \frac{\pi_\theta(o_i |q)}{\pi_{\theta_{old}}(o_i |q)} A_i, \text{clip} \left( \frac{\pi_\theta(o_i |q)}{\pi_{\theta_{old}}(o_i |q)}, 1 - \epsilon, 1 + \epsilon \right)  A_i \right) - \beta \mathbb{D}_{KL}\left(\pi_{\theta} || \pi_{ref}\right)\right) ,
\end{split}
\label{eq:GRPO-obj}
\end{equation}
\begin{equation}
    \mathbb{D}_{KL}\left(\pi_{\theta} || \pi_{ref}\right) = \frac{\pi_{ref}(o_i|q)}{\pi_{\theta}(o_i|q)}- \log\frac{\pi_{ref}(o_i|q)}{\pi_{\theta}(o_i|q)} - 1,
\end{equation}
\begin{equation}
    A_i = \frac{r_i - {\mathrm mean(\{r_1, r_2, \cdots, r_G\})}}{{\mathrm std(\{r_1, r_2, \cdots, r_G\})}}.
\label{eq:Adv}
\end{equation}

\subsection{Training Template}

\begin{figure}[htbp]
    \centering
    \includegraphics[width=\textwidth]{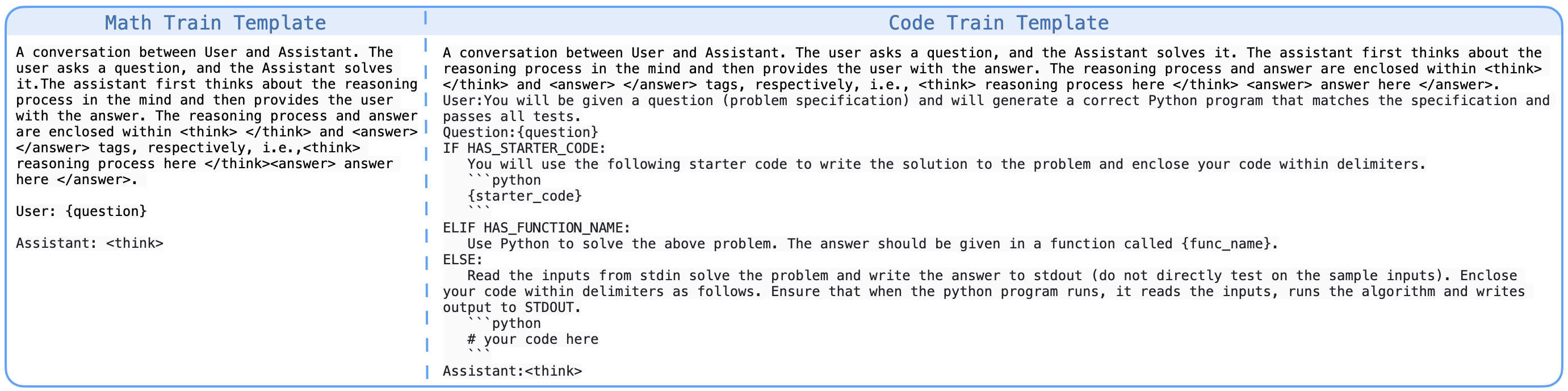}
    \caption{Math and Code training templates.}
    \label{fig:training template}
\end{figure}

For reproducibility, we adopt the training template specified in DeepSeek-R1-Zero, as shown in Figure \ref{fig:training template}. This template provides neither explicit reflective reasoning cues nor specific solution strategies, ensuring that the model relies solely on its intrinsic reasoning capabilities to solve problems. 

For code questions that include starter code, we incorporate the problem description and the function name provided by the starter into the corresponding question. 

\newpage
\section{Approach}
\label{gen_inst}

\subsection{Challenges with Vanilla GRPO Implementation}
In our initial experiments using vanilla GRPO to train the Qwen-2.5-32B-Base \cite{yang2024qwen2} base model, we encountered several training bottlenecks that prevented the model from reaching the desired R1-Zero performance on both math and code benchmarks. These challenges include:

\begin{enumerate}
    \item \textbf{Intrinsic Response-Length Conflict between Math and Code:} We find that math problems often elicit longer and meticulous step-by-step reasoning patterns (long CoT), while RL training on code data does not encourage the model to generate longer responses. Directly mixing these two types of data creates an intrinsic conflict and limits the model's ability to optimally develop either the necessary verbosity required for mathematical reasoning or the succinct conventions preferred in coding tasks. Consequently, this conflict results in suboptimal performance in both domains.
    
    \item \textbf{Degraded Training Efficiency due to Identical Group Rewards:} The GRPO algorithm relies on non-zero reward variance within a sampled group to compute the advantage (Eq.~\ref{eq:Adv}). When all or most rollouts within a group yield identical reward values, the computed advantage approaches zero. Consequently, when a large part of the training batch exhibits this phenomenon, the effective gradient contribution becomes minimal, significantly hampering training efficiency.
    
    \item \textbf{Premature Performance Saturation:} GRPO training often encounters performance plateaus in evaluation benchmarks, characterized by an early slowdown in reward improvement. This issue typically arises from an inadequate quality of the data set. Specifically, when training data lacks sufficient complexity or diversity, particularly if overly simple problems dominate, the model tends to conservatively maintain its performance in these easier tasks. Consequently, it struggles to engage in the deeper and extended reasoning processes required for more challenging problems, limiting the overall development of its reasoning capabilities.
\end{enumerate}

\subsection{Two Stage Training}
To address the challenge of the intrinsic response-length conflict between math and code, we propose a two-stage training paradigm:

\begin{enumerate}
    \item \textbf{Stage 1 (Eliciting Reasoning Abilities):} The initial training phase focuses solely on challenging mathematical data. The goal of this stage is to encourage the development of extended CoT capabilities, including reflective pauses, backtracking behaviors, and step-by-step decomposition.
    
    \item \textbf{Stage 2 (Skill Integration):} In this stage, the coding data is introduced into the training process. Using the reasoning foundation established in Stage 1, this stage aims at gaining proficiency in code generation. Beyond directly enhancing the coding capabilities, this stage also aims to elicit procedural thinking, recursion, and tool-calling capabilities from the model.
\end{enumerate}

\begin{figure}[H]
    \centering
    \includegraphics[width=0.8\textwidth]{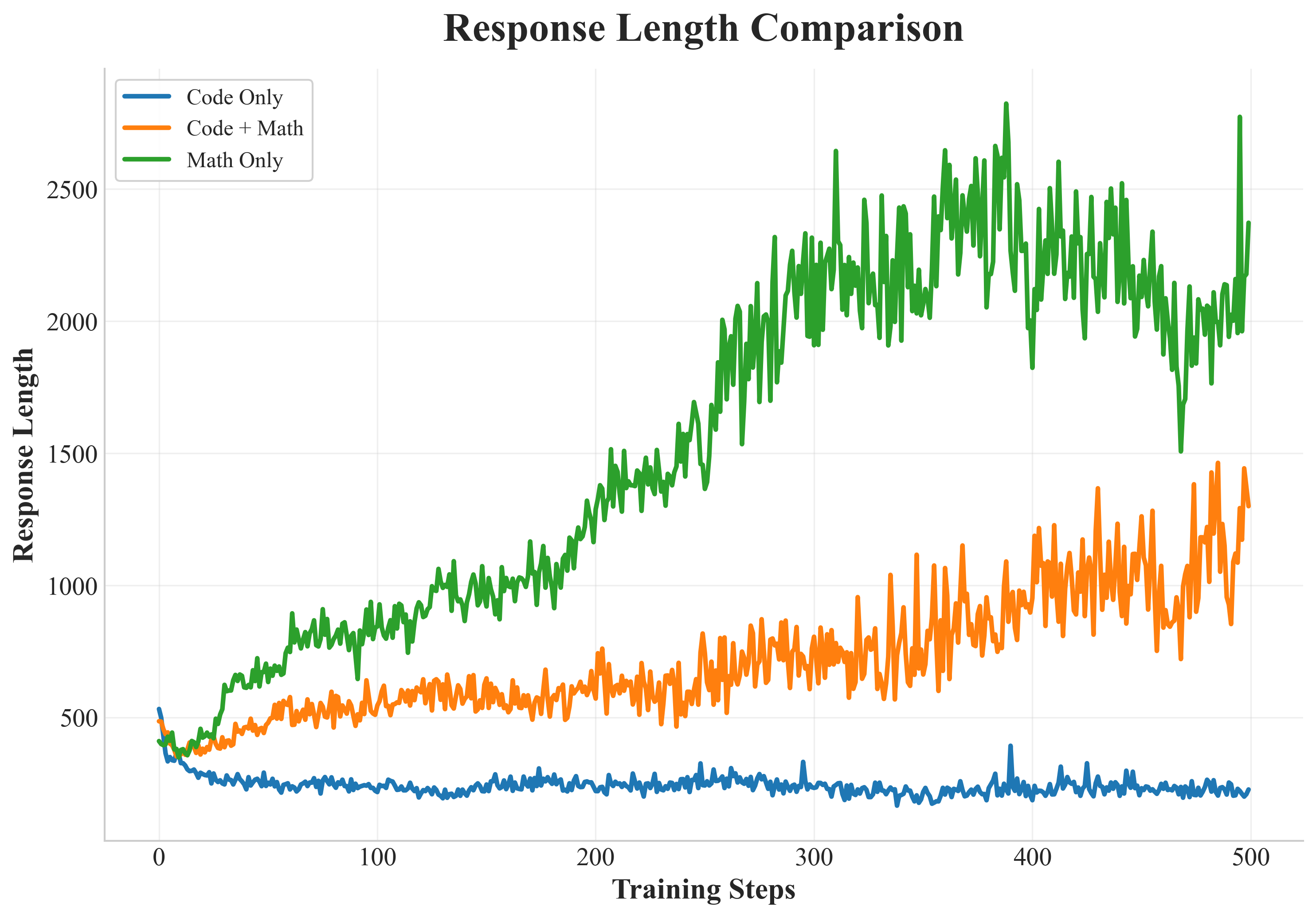}
    \caption{Response length of different training data strategies.}
    \label{fig:2 stage}
\end{figure}

\subsubsection{Comparative Analysis of Training Strategies}
Figure \ref{fig:2 stage} illustrates the response length growth under different training strategies. To validate our staged training approach, we conduct comparative studies analyzing reasoning behaviors and performance across different training paradigms.

\paragraph{Mixed Training} Models trained directly on a mix of math and code data exhibit limited growth in response length and inferior benchmark performance. Case studies reveal inconsistent reasoning patterns: While mathematical problems elicit some reasoning patterns, code problems frequently produce short, direct responses focused primarily on immediate code output with minimal preliminary analysis or planning.
\begin{figure}[H]
    \centering
    \includegraphics[width=\textwidth]{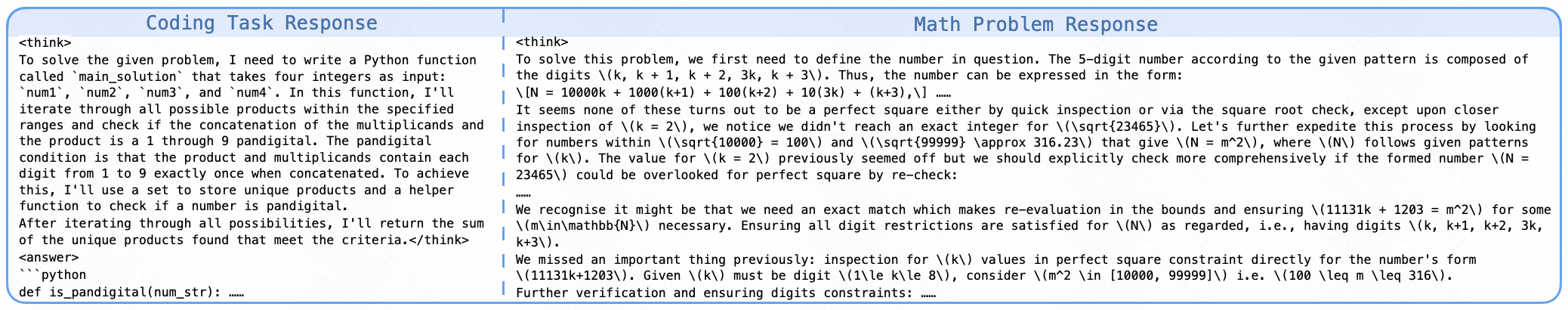}
    \caption{Case of Mixed Training.}
    \label{fig:case1}
\end{figure}
\paragraph{Math-Only Training} Training only on math data led to a consistent increase in response length and strong performance on math benchmarks. Importantly, this fosters robust reasoning abilities that generalize well; when presented with coding tasks, the model attempts detailed step-by-step reasoning. Observed behaviors include meticulous step checking and re-examination within mathematical problem-solving processes. This reflects the ability of mathematical data to elicit reasoning behaviors.
\begin{figure}[H]
    \centering
    \includegraphics[width=\textwidth]{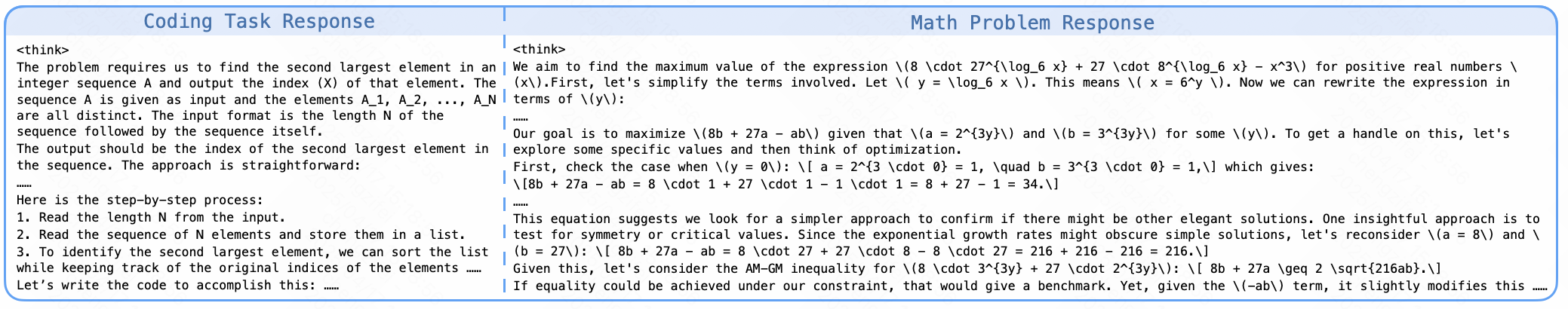}
    \caption{Case of Math-Only Training.}
    \label{fig:case2}
\end{figure}
\paragraph{Code-Only Training} Although performance on code benchmarks improves, there is minimal development of explicit reasoning behaviors, and achieving significant increases in response length proves difficult. Compared to math-only training, responses to both code and math problems are significantly shorter, with solutions to coding tasks often generated directly, lacking substantial step-by-step reasoning or preliminary analysis.
\begin{figure}[H]
    \centering
    \includegraphics[width=\textwidth]{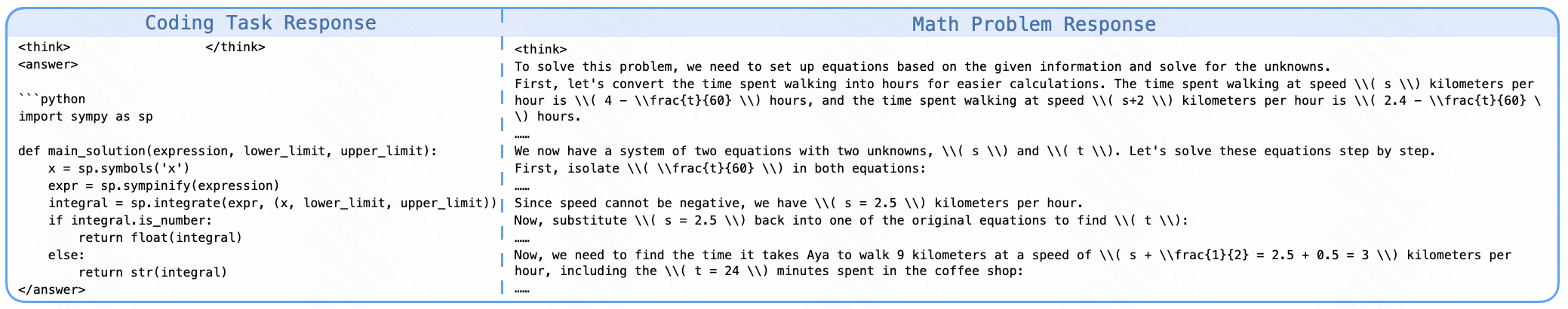}
    \caption{Case of Code-Only Training.}
    \label{fig:case3}
\end{figure}

\paragraph{Staged Training} The proposed two-stage training demonstrates superior results in both the mathematical and coding domains. The model consistently generates detailed step-by-step reasoning patterns when solving mathematical problems, as well as structured reasoning patterns for programming tasks. In particular, sophisticated behaviors emerge, such as models that spontaneously leverage code generation as an auxiliary tool to assist mathematical reasoning. A more detailed analysis of these response patterns is presented in Section \ref{sec:think}. 
\begin{figure}[H]
    \centering
    \includegraphics[width=\textwidth]{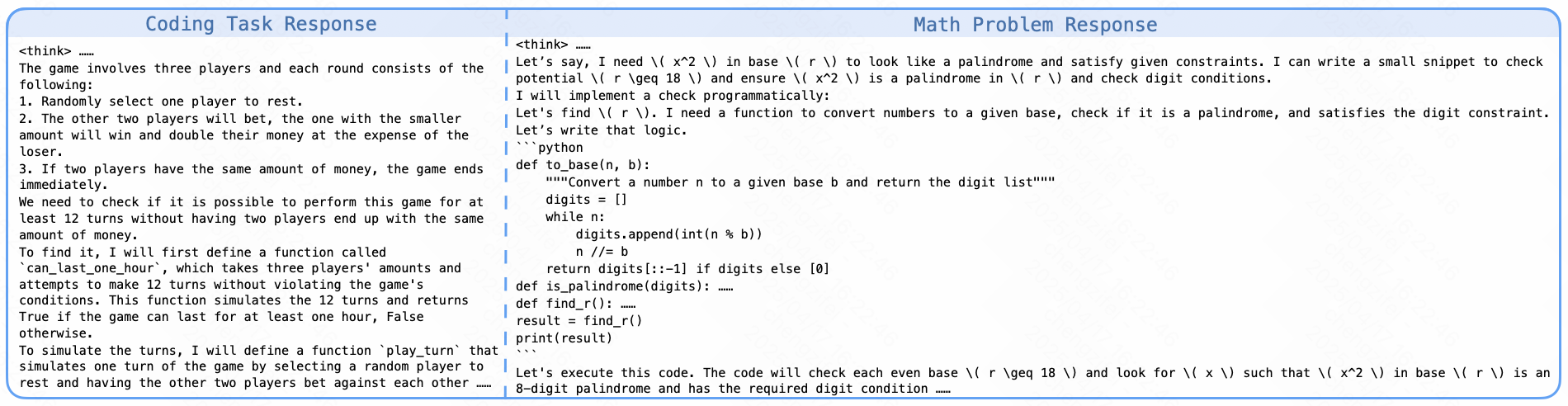}
    \caption{Case of Staged Training.}
    \label{fig:case4}
\end{figure}

\begin{table}[H]
    \centering
    \begin{tabular}{ccc}
        \toprule
        Training Strategy (w/o HR) & AIME24 (Pass@1) & LiveCodeBench (Pass@1) \\
        \midrule
        Naive Mixed Training & 40.5 & 35.1 \\
        Staged Training & 44.3 & 38.7 \\
        \bottomrule \hline
    \end{tabular}
    \caption{Performance comparison of training strategies (w/o History Resampling).}
    \label{tab:training_strategies}
\end{table}

\subsection{History Resampling (HR)}

\begin{figure}[H]
    \centering
    \includegraphics[width=0.7\textwidth]{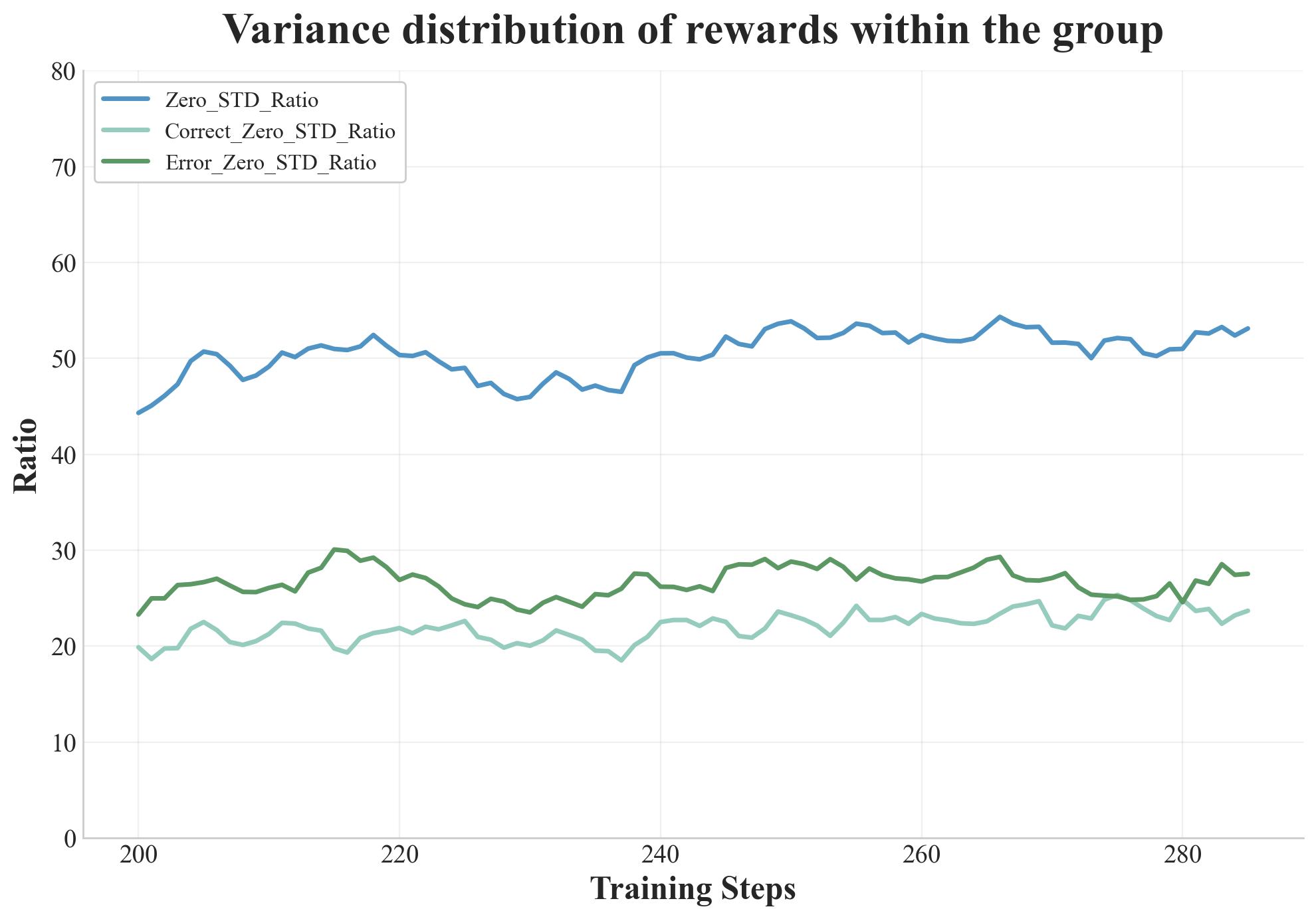}
    \caption{Nearly 50\% of advantages within batches are zero during training (blue line).}
    \label{fig:hr_var}
\end{figure}

During training, nearly 50\% of the sampled groups within batches produce identical rewards (Figure \ref{fig:hr_var}). This scenario typically occurs when the model consistently succeeds on easier problems, leading to minimal variance in rewards and ineffective gradient updates.

To address this inefficiency and enhance the quality of gradient signals, we introduce History Resampling (HR), an epoch-level resampling mechanism. During training, we record the outcomes of all rollout rewards within each epoch. At the end of an epoch, we reconstruct the dataset for the subsequent epoch as follows:
\begin{itemize}
    \item \textbf{Filter Out ''Too Easy'' :} Samples where all rollouts get the correct answer are excluded, as they provide no informative contrastive signals for policy improvement.
    \item \textbf{Retain ''Informative'':} Samples that yield either mixed outcomes (both correct and incorrect rollouts) or exclusively incorrect outcomes are retained. These samples exhibit positive reward variance, ensuring nonzero advantages and effective gradient signals. Additionally, for hard samples where all rollouts were incorrect within the current epoch, we keep them in the dataset. The rationale is that some of these initially challenging problems may become relatively easier for the updated policy, thereby generating positive advantages in subsequent rollouts. The underlying idea of this strategy aligns with curriculum learning\cite{narvekar2020curriculum}, progressively exposing the model to samples that are, on average, more challenging, to improve training efficiency.
\end{itemize}

\begin{figure}[htbp]
    \centering
    \begin{minipage}[b]{0.49\textwidth}
        \centering
        \includegraphics[width=\textwidth]{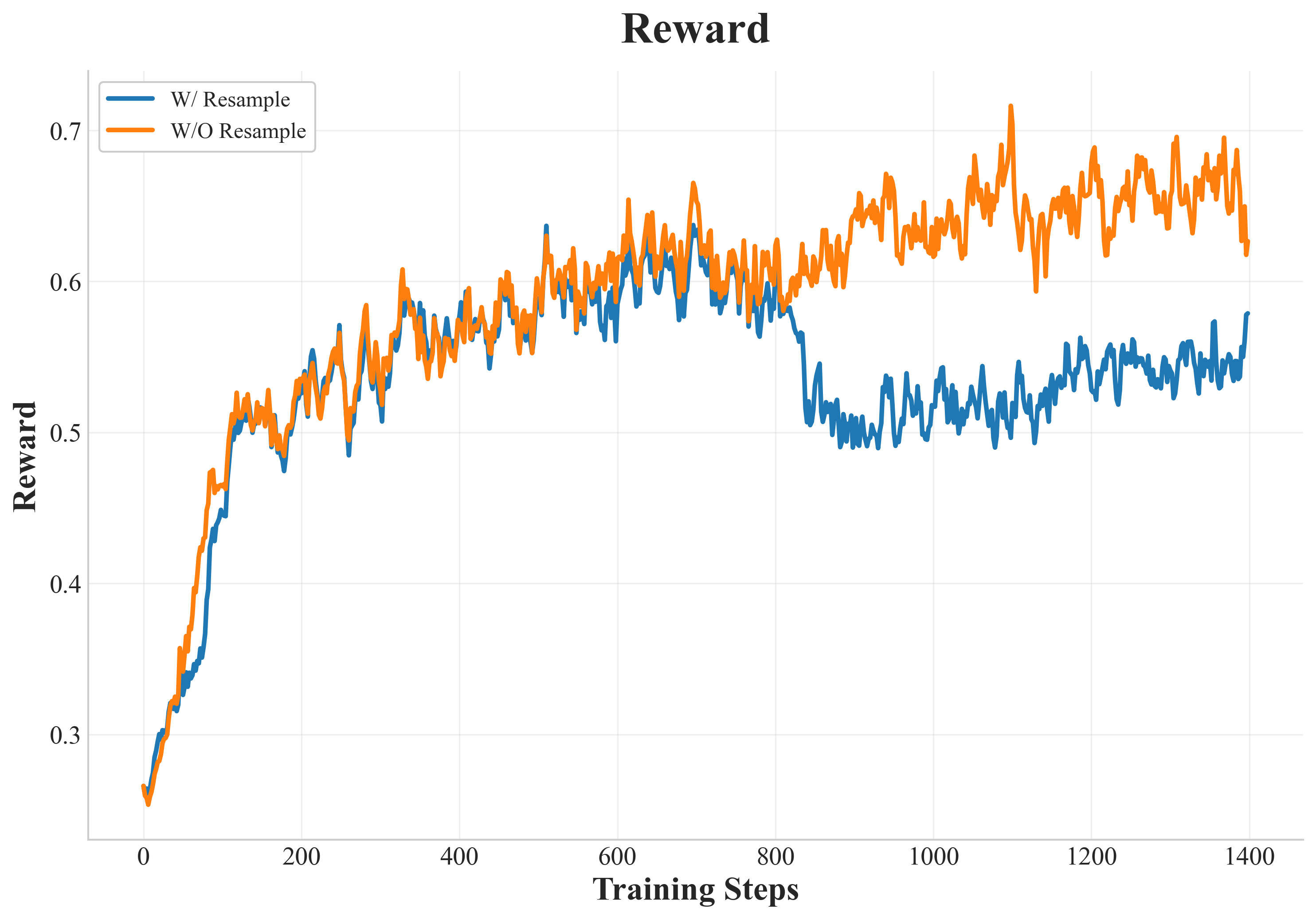}
        \caption*{(a) Reward statistics}
    \end{minipage}
    \hfill
    \begin{minipage}[b]{0.49\textwidth}
        \centering
        \includegraphics[width=\textwidth]{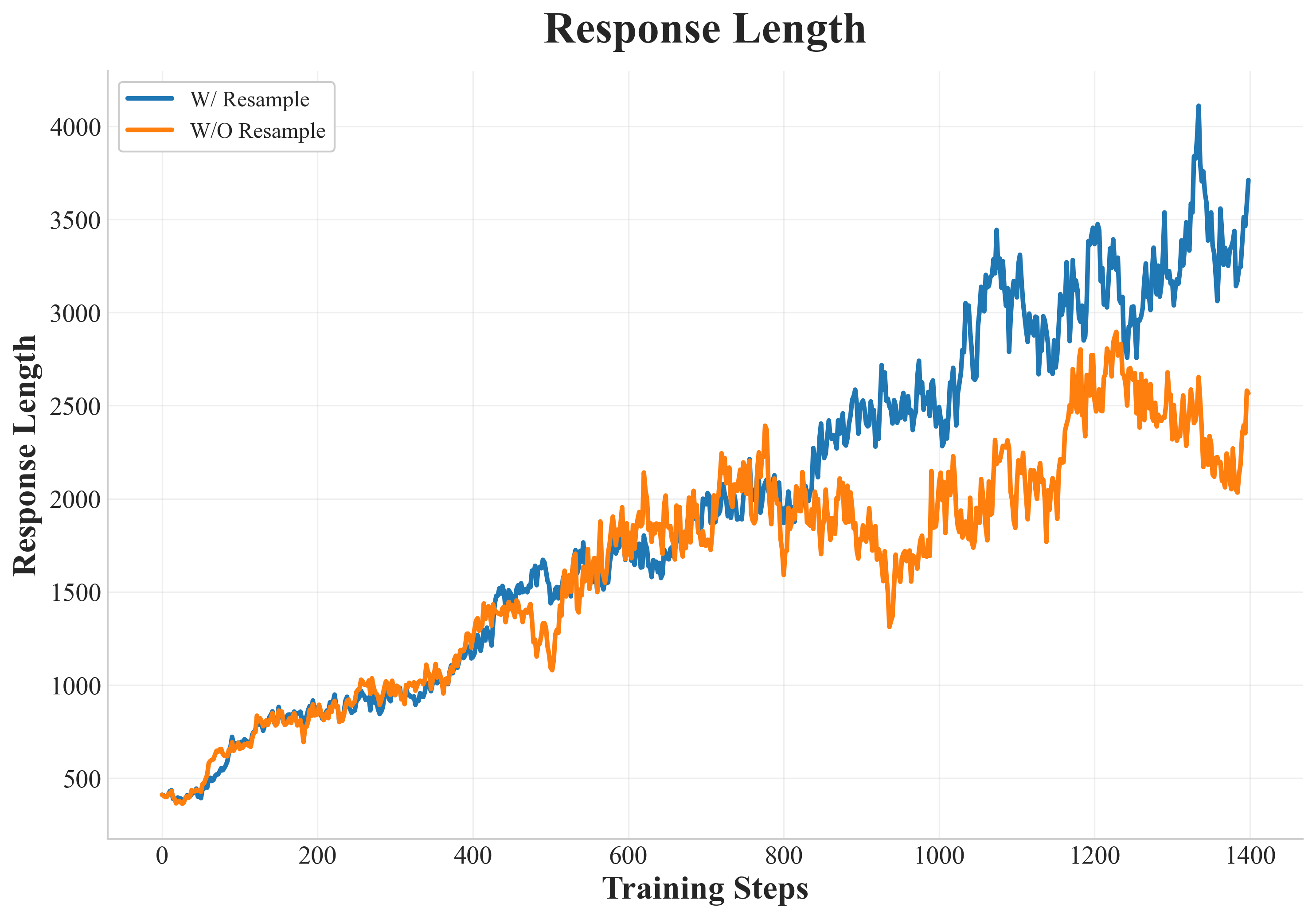}
        \caption*{(b) Length statistics}
    \end{minipage}
    \caption{Training statistics of History Resampling (w/o 2-stage training).}
    \label{fig:hr}
\end{figure}
Compared to the dynamic sampling method\cite{yu2025dapo}, which adjusts the rollout number on the fly, history sampling significantly improves computational efficiency. The positive impact of history resampling on the growth of response length is illustrated in Figure \ref{fig:hr}.

\subsection{Data Curation}
\begin{figure}[htbp]
    \centering
    \includegraphics[width=\textwidth]{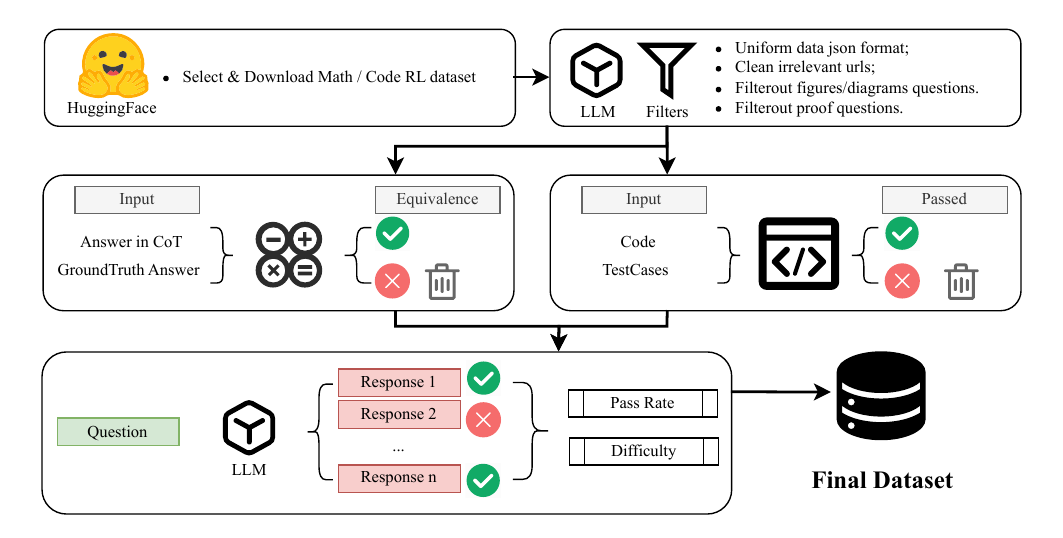}
    \caption{Data curation pipeline of SRPO.}
    \label{fig:data}
\end{figure}
Our data curation pipeline is illustrated in Figure \ref{fig:data}. In the download stage, we select and download math and code datasets relevant to RL training from the HuggingFace open source community. In the cleaning stage, datasets are first transformed into the required JSON format for training. Irrelevant URLs, which commonly appear in samples, are heuristically cleaned. Since certain samples may be difficult to verify or even unsolvable for the model, we employ heuristic rules and LLM-based methods to filter them out. Specifically, for math samples, we remove those containing multiple questions, pure proof-based problems, or those requiring images or tables. For code samples, we exclude problems that depend on specific environments or involve file I/O and network interactions.

In the verification stage, we ensure the correctness and clarity of the mathematical and code questions. The generated solutions are extracted from the CoT of the model. Code solutions are executed within a sandbox environment against a set of test cases (up to 10). Mathematical solutions are validated using an equivalence verifier. Samples that fail this verification process are filtered out. In addition, we assess the intelligibility of answers, eliminating questions with incorrect solutions or ambiguities. Finally, we categorize the verified problems into three difficulty levels: easy, medium, and difficult, based on their observed pass rate (pass@1) and success rates across multiple attempts (pass@k).

\subsection{Reward Design}
To mitigate reward hacking issues caused by the reward model\cite{amodei2016concrete, everitt2017reinforcement, krakovna2020specification, everitt2021reward, gao2022scaling, weng2023reward}, we adopt a rule-based reward system design similar to DeepSeek-R1, dividing the final reward into format reward and accuracy reward.
For the format reward $R_{format}$, we require that the model's final answer strictly adheres to the format "<output>answer</output>". Adhering to the format earns 0.2 reward; otherwise, 0. Additionally, any instance of code-switching or mixing multiple languages in the model’s response will result in a penalty term: \( {Penalty}_{\text{mix}} = -0.1 \).

For the accuracy reward, we evaluate the mathematical and coding tasks separately. In mathematical tasks, we verify the correctness using a mathematical verification tool. If the answer is entirely correct and equivalent, we award a full score of 1 point. In cases where the answer is partially correct (e.g., a multiple-choice question where the model selects only one correct option), we grant a partial score of 0.2 points. Completely incorrect answers receive 0 points.
\[
R_{\text{math}} = 
\begin{cases} 
1, & \text{correct} \\
0.2, & \text{partially correct} \\
0, & \text{otherwise}
\end{cases}
\]

For coding tasks, the model receives 1 point if all test cases pass successfully; if any test case fails, the score awarded is 0 points. 
\[
R_{\text{code}} = 
\begin{cases} 
1, & \text{all test cases pass} \\
0, & \text{otherwise}
\end{cases}
\]

The total reward R is the combined result of the format reward, accuracy reward, and negative penalty:
\[
R = R_{\text{math}/\text{code}} + R_{\text{format}} + {Penalty}_{\text{mix}}
\]

\section{Experiments}
\label{headings}

In this section, we present detailed experimental results for SRPO. We first outline training configurations and key hyperparameters, clarifying our use of popular optimization techniques. Next, we provide evaluation results and settings. We then examine training dynamics, highlighting metrics such as reward trends and response length. Finally, we analyze the model’s thinking behaviors.
\subsection{Training Settings}

We use Qwen-2.5-Base-32B as the initial checkpoint and train with a constant learning rate of 1e-6, using the AdamW optimizer\cite{loshchilov2018adam} ($\beta = [0.9, 0.95]$) without weight decay. We employ vLLM\cite{kwon2023vllm} as our rollout inference framework, sampling 256 prompts per step with 32 rollouts per prompt, with minibatch size of 8,192 for strict on-policy RL training. Stage 1 training runs for 840 steps, followed by Stage 2. After each epoch, we perform history resampling to remove overly simple data.

Specifically, we apply a token-level loss and set the advantage of responses exceeding the maximum response length (10,000 tokens) to zero to prevent length bias from clipping. Additionally, considering the significant distribution difference between the reasoning patterns and the base model, we remove the KL term from the GRPO loss function to encourage exploration.

\subsection{Main Results}

Our approach ultimately achieves 50.0 pass@1 on AIME 24 and 41.6 pass@1 on LiveCodeBench\cite{jain2024livecodebench} (2024-08 – 2025-01, same version as used in DeepSeek's report\cite{deepseek2025r1}), surpassing the multi-domain SOTA performance of DeepSeek-R1-Zero-Qwen-32B and matching DAPO, current-best GRPO-based method, on the AIME24 benchmark.

\begin{table}[H]
    \centering
    \begin{tabular}{ccc}
        \toprule
        Model & AIME24 & LiveCodeBench \\
        \midrule
        DeepSeek-R1-Zero-Qwen-32B & 47.0 & 40.2 \\
        SRPO (Ours) & 50.0 & 41.6 \\
        \bottomrule
    \end{tabular}
    \caption{Performance of SRPO on AIME24 and LiveCodeBench.}
    \label{tab:result}
\end{table}
\subsection{Training Dynamics}
\begin{figure}[htbp]
    \centering
    \begin{minipage}[b]{0.49\textwidth}
        \centering
        \includegraphics[width=\textwidth]{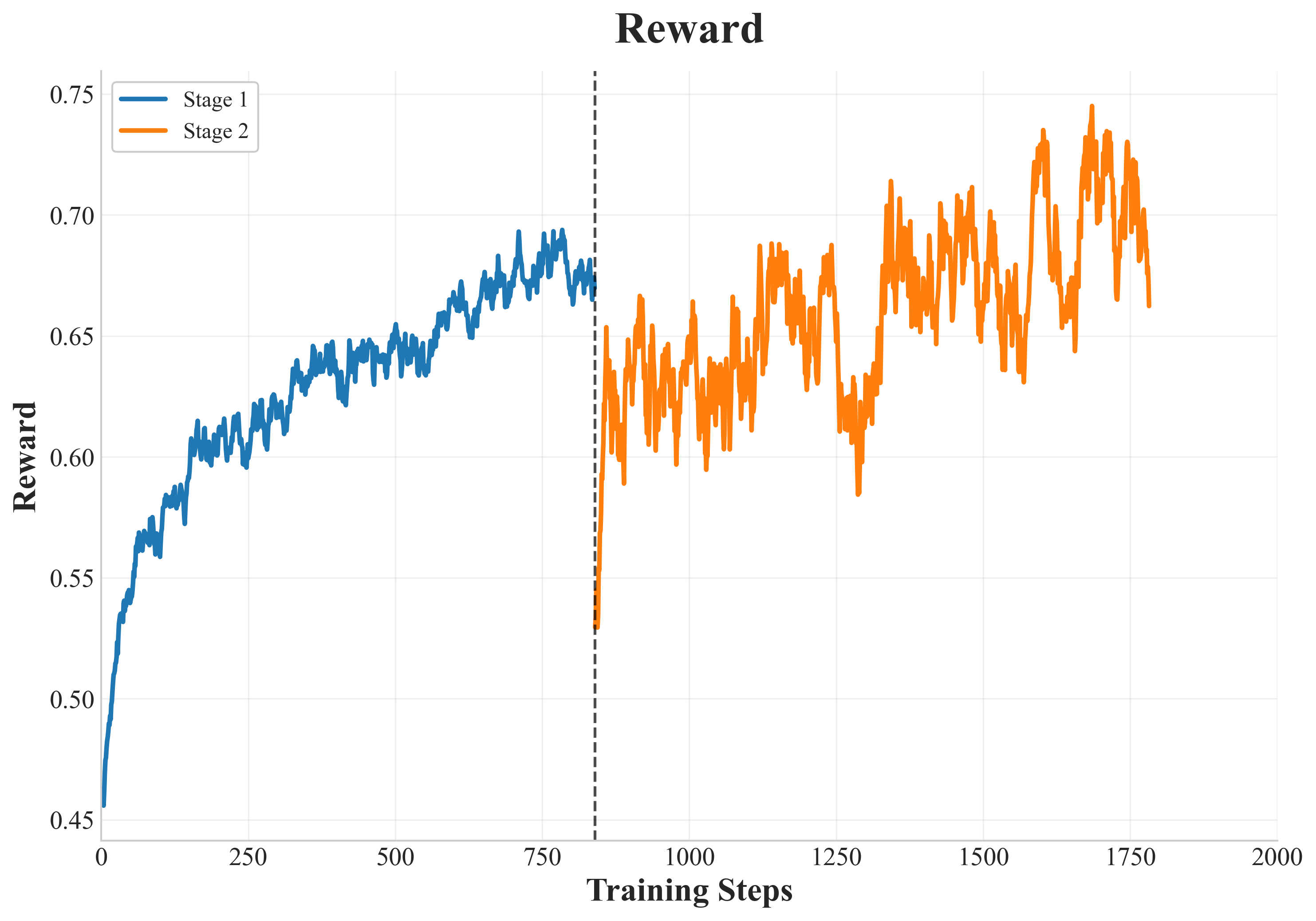}
    \end{minipage}
    \hfill
    \begin{minipage}[b]{0.49\textwidth}
        \centering
        \includegraphics[width=\textwidth]{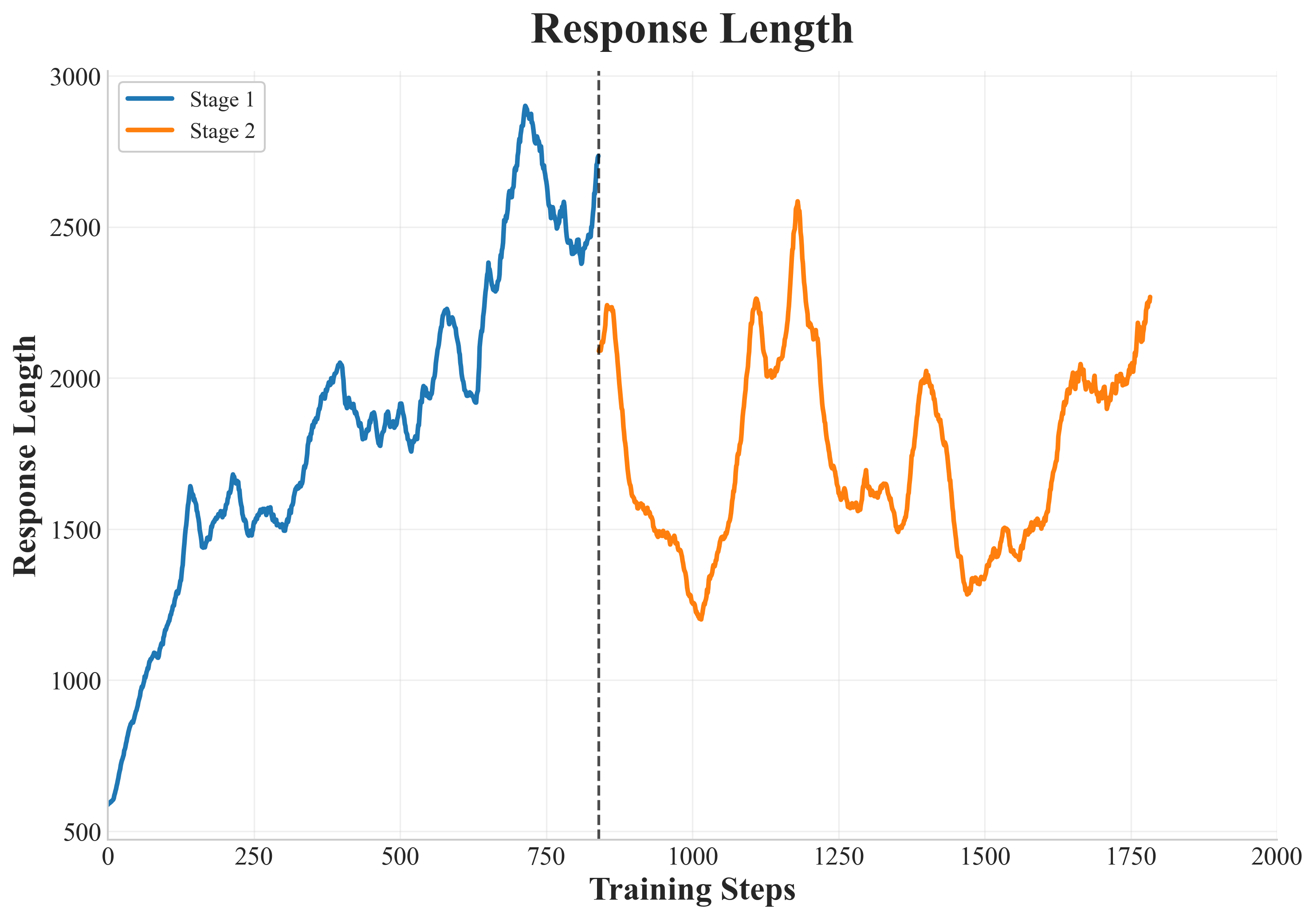}
    \end{minipage}
    \caption{Training dynamics of SRPO.}
    \label{fig:td}
\end{figure}

Figure \ref{fig:td} illustrates the full reward curve and the response length curve of the SRPO training. We proceeded to stage 2 training after the reward growth began to plateau. At the beginning of stage 2, the overall reward decreased due to the model's previously untrained coding capability, after which subsequent training led to a steady increase in reward. After incorporating coding data, the response length does not increase significantly, which is consistent with our expectations, since coding tasks typically do not necessitate longer reasoning chains. Meanwhile, benchmark results indicate consistent and steady improvement in both mathematical and coding abilities of the model, demonstrating the effectiveness of our approach.

In particular, the proposed history resampling technique ensures consistently effective gradient updates at each training step, directly increasing the proportion of informative gradients. This improved sample efficiency results in steady reward growth, clearly demonstrating the enhanced training efficiency achieved by our resampling strategy. 

\subsection{Thinking Behavior Analysis}
\label{sec:think}
We identify three groups of representative reflection patterns following a methodology similar to \cite{openai2024o1, yeo2025demystifying}. Those groups include recheck (e.g. "recheck", "reevaluate", "reexamine", "rethink", "double check"), hesitation ("wait", "but", "maybe", "aha") and explore ("another way", "another approach", "another method", "but how", "hold on"). We count the number of responses containing any of these patterns as 'aha responses', and identify the average response length (both aha and non-aha responses). 

We are surprised to find that in the course of RL training, the frequency of self-reflection, correction, and backtracking of the model gradually increases. This demonstrates that the model develops "self-verification" ability. We believe that the emergence of "reflection" analogous to human cognitive processes represents an adaptive behavior exhibited by the model during policy optimization in RL.

\begin{figure}[htbp]
    \centering
    \includegraphics[width=\textwidth]{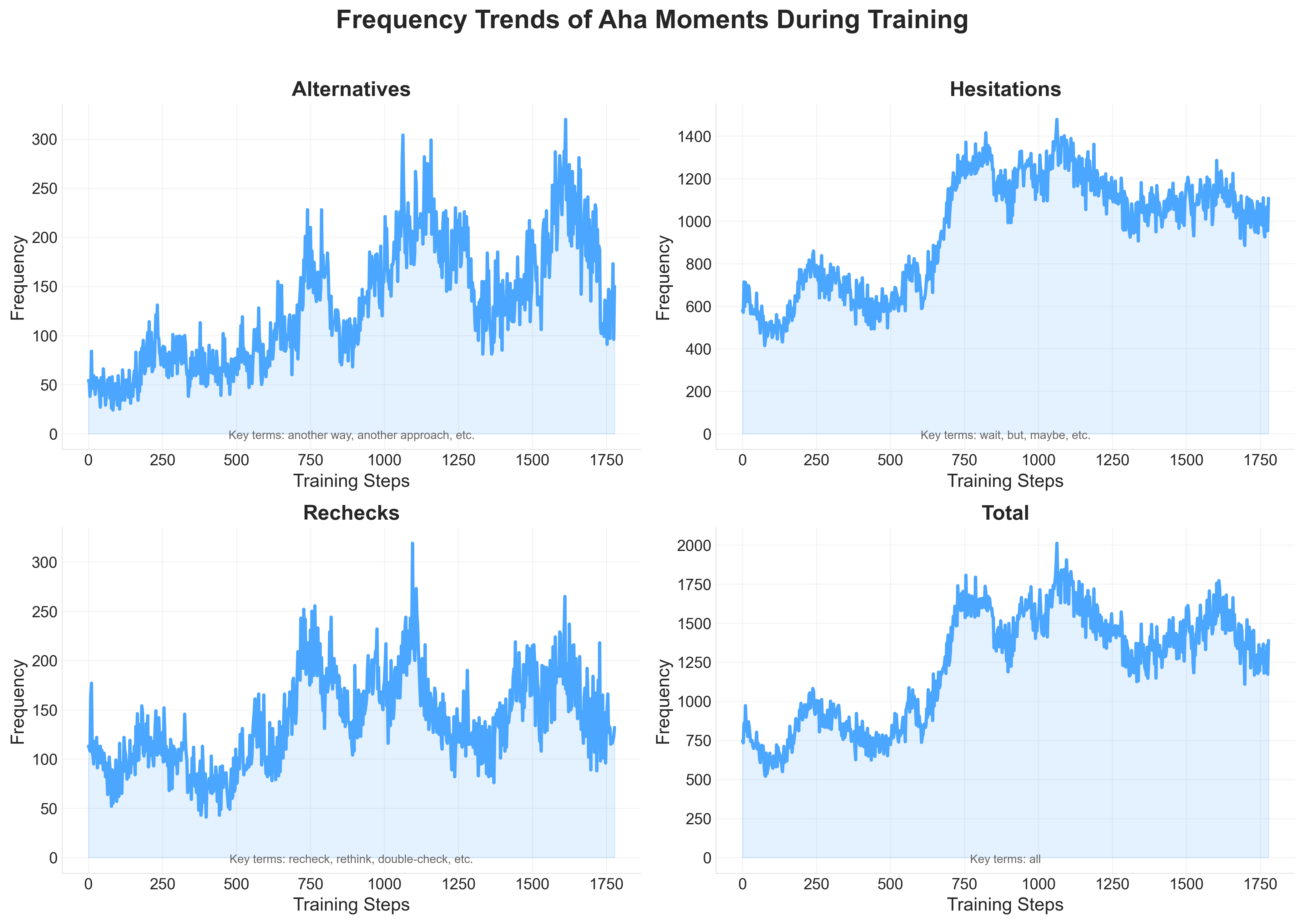}
    \caption{Occurrence of various reasoning patterns during training.}
    \label{fig:aha_freq}
\end{figure}

\newpage
As shown in Figure \ref{fig:aha_freq}, there is little evidence of checking and reflecting at an early stage of model training. However, as training progresses, the model exhibits clear reflection and backtracking behaviors, forming response modes such as stepwise reasoning, numerical substitution, one-by-one verification, and self-optimization (cases shown in the Appendix).

\begin{figure}[htbp]
    \centering
    \includegraphics[width=\textwidth]{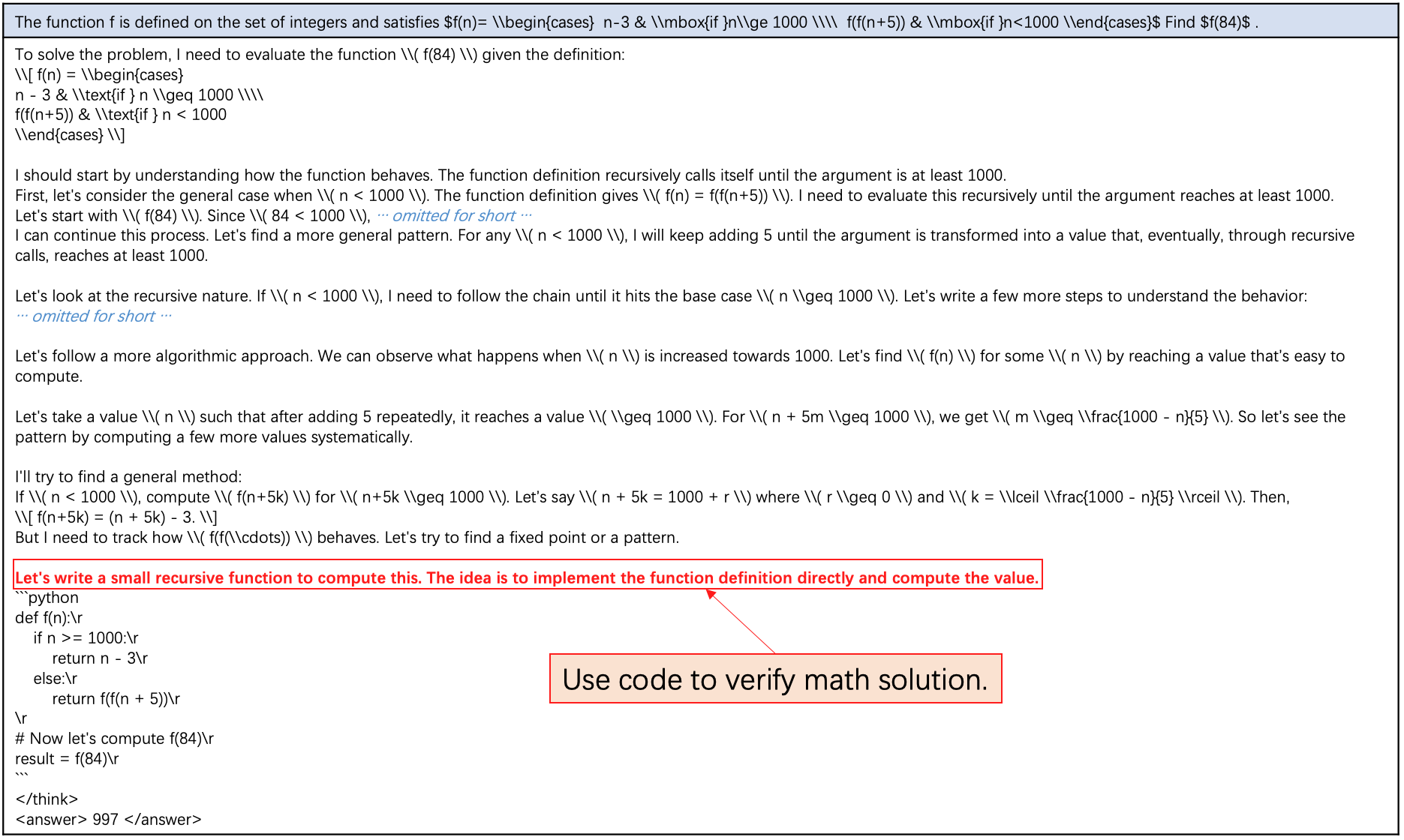}
    \caption{Math solution verification using code.}
    \label{fig:aha_4}
\end{figure}

Another interesting observation is that code is used for verification during mathematical problem solving. For example, in the math example above (Figure \ref{fig:aha_4}), the model first provides the solution process through mathematical reasoning and then spontaneously writes the program code to verify the results of the scheme. As previously noted in the two-stage training section, such cases illustrate how models employ procedural reasoning to self-correct and make subsequent attempts. In addition, the rich world knowledge allows LLMs to solve problems in a variety of ways. Using a combination of methods becomes a natural choice. This phenomenon further indicates that the model has already mastered the skills of thinking broadly and using extensive tools in problem solving.

\section{Conclusion}
In this paper, we introduce two-Staged history-Resampling Policy Optimization  (SRPO), the first implementation to replicate the reasoning performance of DeepSeek-R1-Zero-Qwen-32B on math and code tasks. We propose a two-stage training paradigm, epoch-level history resampling, and a dedicated data curation pipeline tailored for math-code joint RL training. Our detailed experiments and empirical insights offer practical guidance and valuable experience for future research on generalizable large-scale reinforcement learning.

\newpage
{
\setlength{\parindent}{0pt}

\section*{Contributions}

\subsection*{Project Lead}
Xiaojiang Zhang

\subsection*{Algorithm}
Xiaojiang Zhang, Jinghui Wang, Zifei Cheng, Wenhao Zhuang, Zheng Lin, Minglei Zhang

\subsection*{Infrastructure}
Shaojie Wang, Yinghan Cui, Chao Wang, Junyi Peng, Xiaojiang Zhang, Jinghui Wang, Wenhao Zhuang

\subsection*{Dataset}
Jinghui Wang, Wenhao Zhuang, Xiaojiang Zhang, Zheng Lin, Shimiao Jiang, Minglei Zhang, Shiqi Kuang, Shouyu Yin, Chaohang Wen

\subsection*{Supervision}
Haotian Zhang, Bin Chen, Bing Yu
}

\newpage
\bibliography{main}

\begin{thebibliography}{10}

\bibitem{openai2024reason}
OpenAI.
\newblock Learning to reason with llms, 2024.

\bibitem{openai2024o1}
Aaron Jaech, Adam Kalai, Adam Lerer, Adam Richardson, Ahmed El-Kishky, Aiden Low, Alec Helyar, Aleksander Madry, Alex Beutel, Alex Carney, et~al.
\newblock Openai o1 system card.
\newblock {\em arXiv preprint arXiv:2412.16720}, 2024.

\bibitem{deepseek2025r1}
Daya Guo, Dejian Yang, Haowei Zhang, Junxiao Song, Ruoyu Zhang, Runxin Xu, Qihao Zhu, Shirong Ma, Peiyi Wang, Xiao Bi, et~al.
\newblock Deepseek-r1: Incentivizing reasoning capability in llms via reinforcement learning.
\newblock {\em arXiv preprint arXiv:2501.12948}, 2025.

\bibitem{openai2023gpt4}
OpenAI.
\newblock Gpt4 technical report.
\newblock {\em arXiv preprint arXiv:2303.08774}, 2023.

\bibitem{anthropic2024claude}
Anthropic.
\newblock Claude 3.5 sonnet, 2024.

\bibitem{deepseek2024v3}
Aixin Liu, Bei Feng, Bing Xue, Bingxuan Wang, Bochao Wu, Chengda Lu, Chenggang Zhao, Chengqi Deng, Chenyu Zhang, Chong Ruan, et~al.
\newblock Deepseek-v3 technical report.
\newblock {\em arXiv preprint arXiv:2412.19437}, 2024.

\bibitem{xai2024grok}
XAI.
\newblock Grok 3 beta — the age of reasoning agents, 2024.

\bibitem{deepmind2024gemini}
Google DeepMind.
\newblock Gemini 2.0 flash thinking, 2024.

\bibitem{qwen2024qwq}
Qwen.
\newblock Qwq-32b: Embracing the power of reinforcement learning, 2024.

\bibitem{du2025kimi}
Angang Du, Bofei Gao, Bowei Xing, Changjiu Jiang, Cheng Chen, Cheng Li, Chenjun Xiao, Chenzhuang Du, Chonghua Liao, et~al.
\newblock Kimi k1. 5: Scaling reinforcement learning with llms.
\newblock {\em arXiv preprint arXiv:2501.12599}, 2025.

\bibitem{chen2025empirical}
Zhipeng Chen, Yingqian Min, Beichen Zhang, Jie Chen, Jinhao Jiang, Daixuan Cheng, Wayne~Xin Zhao, Zheng Liu, Xu~Miao, Yang Lu, et~al.
\newblock An empirical study on eliciting and improving r1-like reasoning models.
\newblock {\em arXiv preprint arXiv:2503.04548}, 2025.

\bibitem{yu2025dapo}
Qiying Yu, Zheng Zhang, Ruofei Zhu, Yufeng Yuan, Xiaochen Zuo, Yu~Yue, Tiantian Fan, Gaohong Liu, Lingjun Liu, Xin Liu, et~al.
\newblock Dapo: An open-source llm reinforcement learning system at scale.
\newblock {\em arXiv preprint arXiv:2503.14476}, 2025.

\bibitem{shao2024deepseekmath}
Zhihong Shao, Peiyi Wang, Qihao Zhu, Runxin Xu, Junxiao Song, Xiao Bi, Haowei Zhang, Mingchuan Zhang, YK~Li, Y~Wu, et~al.
\newblock Deepseekmath: Pushing the limits of mathematical reasoning in open language models.
\newblock {\em arXiv preprint arXiv:2402.03300}, 2024.

\bibitem{jain2024livecodebench}
Naman Jain, King Han, Alex Gu, Wen-Ding Li, Fanjia Yan, Tianjun Zhang, Sida Wang, Armando Solar-Lezama, Koushik Sen, and Ion Stoica.
\newblock Livecodebench: Holistic and contamination free evaluation of large language models for code.
\newblock {\em CoRR}, 2024.

\bibitem{yang2024qwen2}
An~Yang, Baosong Yang, Beichen Zhang, Binyuan Hui, Bo~Zheng, Bowen Yu, Chengyuan Li, Dayiheng Liu, Fei Huang, Haoran Wei, et~al.
\newblock Qwen2. 5 technical report.
\newblock {\em arXiv preprint arXiv:2412.15115}, 2024.

\bibitem{narvekar2020curriculum}
Sanmit Narvekar, Bei Peng, Matteo Leonetti, Jivko Sinapov, Matthew~E Taylor, and Peter Stone.
\newblock Curriculum learning for reinforcement learning domains: A framework and survey.
\newblock {\em Journal of Machine Learning Research}, 21(181):1--50, 2020.

\bibitem{amodei2016concrete}
Dario Amodei, Chris Olah, Jacob Steinhardt, Paul Christiano, John Schulman, and Dan Mané.
\newblock Concrete problems in {AI} safety.
\newblock {\em arXiv preprint arXiv:1606.06565}, 2016.

\bibitem{everitt2017reinforcement}
Tom Everitt, Victoria Krakovna, Laurent Orseau, Marcus Hutter, and Shane Legg.
\newblock Reinforcement learning with a corrupted reward channel.
\newblock {\em arXiv preprint arXiv:1705.08417}, 2017.

\bibitem{krakovna2020specification}
Victoria Krakovna, Jonathan Uesato, Vladimir Mikulik, Matthew Rahtz, Tom Everitt, Ramana Kumar, Zac Kenton, Jan Leike, and Shane Legg.
\newblock Specification gaming: the flip side of {AI} ingenuity.
\newblock {\em arXiv preprint arXiv:2002.04871}, 2020.

\bibitem{everitt2021reward}
Tom Everitt, Marcus Hutter, Ramana Kumar, and Victoria Krakovna.
\newblock Reward tampering problems and solutions in reinforcement learning: A causal influence diagram perspective.
\newblock {\em arXiv preprint arXiv:2108.08901}, 2021.

\bibitem{gao2022scaling}
Leo Gao, John Schulman, and Jacob Hilton.
\newblock Scaling laws for reward model overoptimization.
\newblock {\em arXiv preprint arXiv:2210.01753}, 2022.

\bibitem{weng2023reward}
Lillian Weng.
\newblock Reward hacking in reinforcement learning.
\newblock \url{lilianweng.github.io}, 2023.
\newblock Accessed: 2025-04-18.

\bibitem{loshchilov2018adam}
Ilya Loshchilov and Frank Hutter.
\newblock Decoupled weight decay regularization.
\newblock In {\em International Conference on Learning Representations}, 2019.

\bibitem{kwon2023vllm}
Woosuk Kwon, Zhuohan Li, Siyuan Zhuang, Ying Sheng, Lianmin Zheng, Cody~Hao Yu, Joseph~E. Gonzalez, Hao Zhang, and Ion Stoica.
\newblock Efficient memory management for large language model serving with pagedattention, 2023.

\bibitem{yeo2025demystifying}
Edward Yeo, Yuxuan Tong, Morry Niu, Graham Neubig, and Xiang Yue.
\newblock Demystifying long chain-of-thought reasoning in llms.
\newblock {\em arXiv preprint arXiv:2502.03373}, 2025.

\end{thebibliography}
\newpage
\appendix

\section{More Show Cases of the Thinking Behavior}

\begin{figure}[H]
    \centering
    \includegraphics[width=\textwidth]{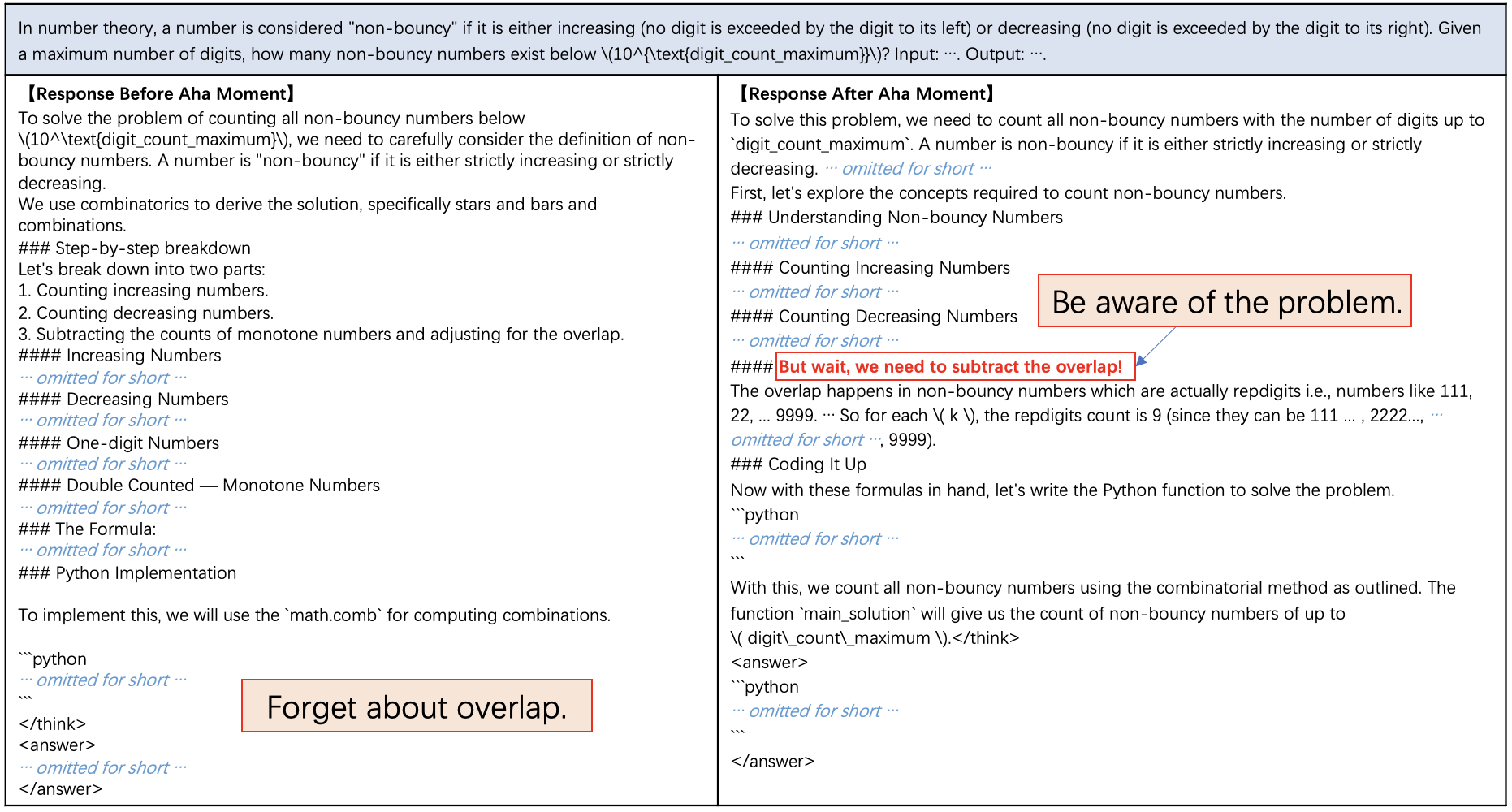}
    \caption{Response with self-correction}
    \label{fig:aha_1}
\end{figure}
\begin{figure}[H]
    \centering
    \includegraphics[width=\textwidth]{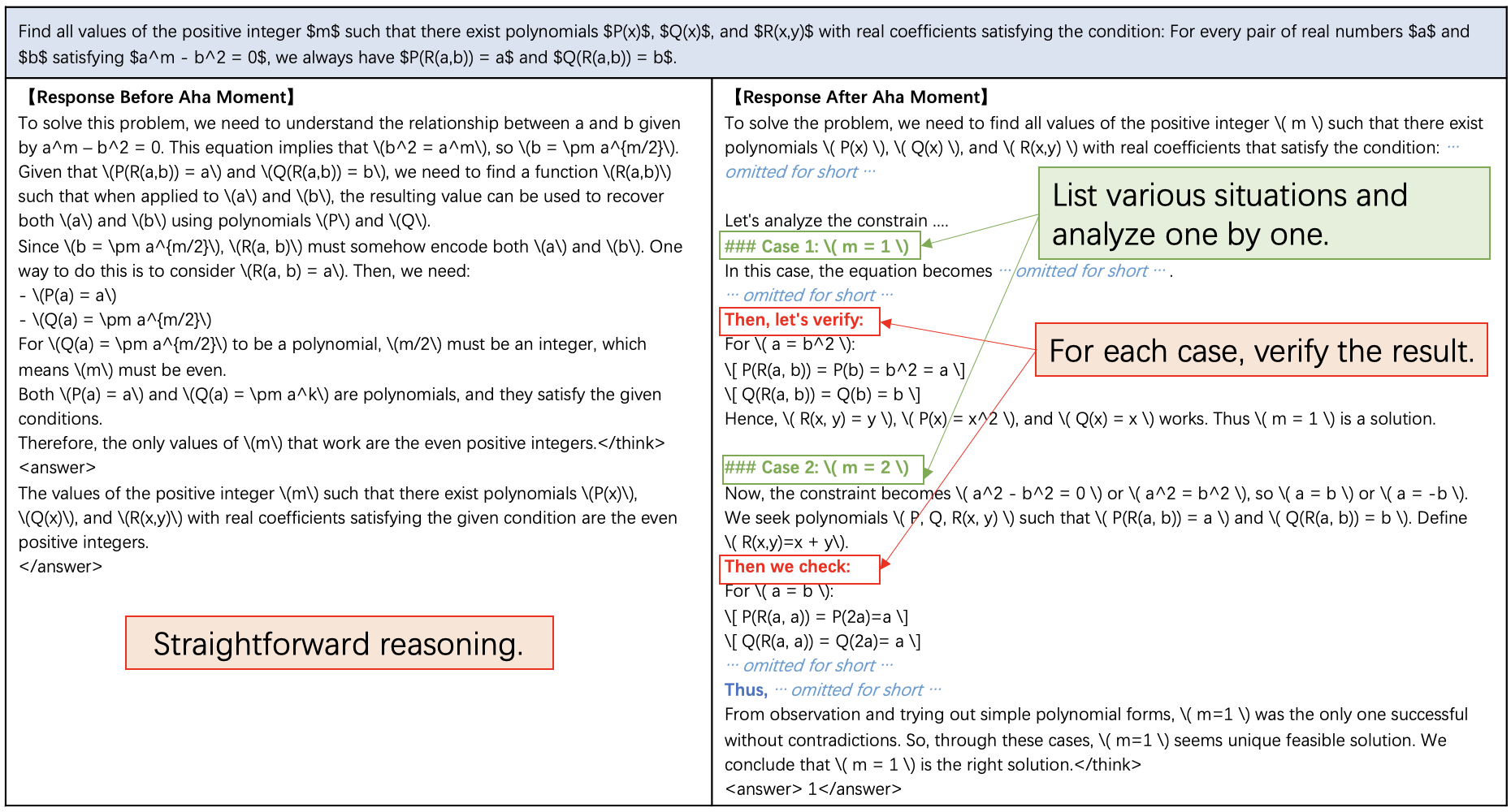}
    \caption{Response with numerical substitution (in green) and verification one by one (in red)}
    \label{fig:aha_2}
\end{figure}
\begin{figure}[H]
    \centering
    \includegraphics[width=\textwidth]{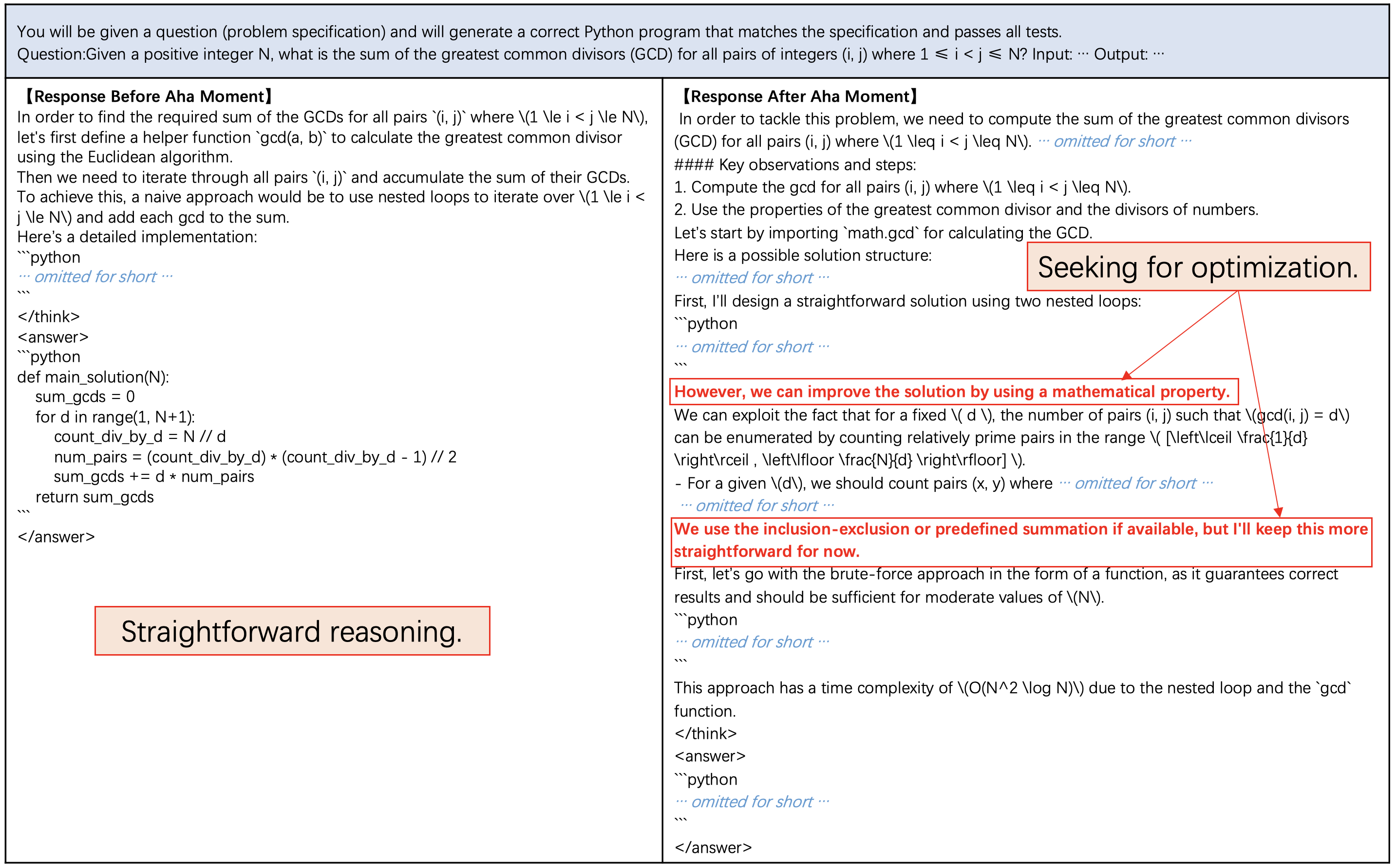}
    \caption{Response with self-optimazition}
    \label{fig:aha_3}
\end{figure}

\end{document}